\documentclass[11pt]{article}
\usepackage[preprint]{acl}
\usepackage{graphicx}
\usepackage{booktabs}
\usepackage{amsmath}

\title{Tokka-Bench: Evaluating Tokenizers Across\\100 Natural and 20 Programming Languages}

\author{Ben Gubler \\
  \texttt{github.com/bgub/tokka-bench} \\
  \texttt{tokka-bench.streamlit.app}}

\begin{document}
\maketitle

\begin{abstract}
Large language models rely on subword tokenizers whose quality varies across languages, yet no standardized multi-metric framework exists for broad comparative evaluation.
We introduce \textsc{Tokka-Bench}, an open-source framework that evaluates tokenizers on five complementary metrics---bytes per token, unique token coverage, subword fertility, word-split rate, and vocabulary composition---across 100 natural languages (30+ scripts) and 20 programming languages, using language-aware segmentation adapted to each writing system.
Comparing seven BPE tokenizers (GPT-2, GPT-4, gpt-oss, Llama~3.1, Gemma~3, Qwen3, and Kimi~K2) within individual languages, we find that vocabulary allocation strategy matters more than raw vocabulary size, and that programming-language efficiency has converged among recent tokenizers despite divergent natural-language profiles.
The framework, data, and interactive dashboard are publicly available.
\end{abstract}

\section{Introduction}

Tokenization converts raw text into subword units and is the first step in every modern large language model (LLM).
Byte Pair Encoding \citep{sennrich2016bpe} and its SentencePiece variant \citep{kudo2018sentencepiece} dominate, but vocabulary quality varies dramatically across languages.

These disparities have far-reaching consequences.
A tokenizer requiring many tokens for equivalent content reduces effective learning per training step.
At inference time, inefficient tokenization lowers throughput, increases cost, and accelerates context-window exhaustion \citep{ahia2023all,petrov2023language}.

Despite growing awareness, the community lacks a standardized, multi-metric framework for comparing tokenizers across languages.
Prior work has focused on individual metrics such as subword fertility \citep{acs2019bert,rust2021good} or cost parity \citep{ahia2023all}, typically for a narrow set of languages.

We present \textsc{Tokka-Bench}, an open-source framework for reproducible tokenizer comparison.
Our contributions are:

\begin{enumerate}
\item \textbf{A reusable benchmark and tool} covering 100 natural languages (30+ scripts) and 20 programming languages, with an interactive Streamlit dashboard.
\item \textbf{Five complementary metrics} capturing compression efficiency, vocabulary coverage, subword granularity, word-boundary alignment, and vocabulary composition.
\item \textbf{Language-aware segmentation} adapting the definition of a linguistic unit to each writing system (Section~\ref{sec:segmentation}).
\item \textbf{A demonstration analysis} of seven BPE tokenizers across three generations.
\end{enumerate}

\section{Related Work}

\citet{acs2019bert} introduced subword fertility as a metric for comparing BERT tokenizers across languages.
\citet{rust2021good} expanded on this, showing that monolingual performance of multilingual models correlates with tokenizer quality across 22 typologically diverse languages.

\citet{petrov2023language} demonstrated that tokenizer disparities introduce systematic unfairness, with low-resource languages requiring up to 15$\times$ more tokens for equivalent content.
\citet{ahia2023all} quantified the economic implications, showing that speakers of many languages are systematically overcharged by token-based API pricing.

\textsc{Tokka-Bench} extends this line of work with a unified, multi-metric framework covering an order of magnitude more languages, with language-aware segmentation.

\section{Methodology}

\subsection{Datasets}

We draw evaluation text from three sources:
\textbf{FineWeb} \citep{penedo2024fineweb} for English web text,
\textbf{FineWeb-2} for 99 additional natural languages, and
\textbf{StarCoder} \citep{li2023starcoder} for 20 programming languages.
For each language, we sample a contiguous 2\,MB UTF-8 block.

\subsection{Tokenizers Evaluated}

Table~\ref{tab:tokenizers} lists the seven tokenizers, spanning vocabulary sizes from 50K (GPT-2) to 262K (Gemma~3).

\begin{table}[t]
\centering
\small
\begin{tabular}{@{}lrr@{}}
\toprule
\textbf{Tokenizer} & \textbf{Vocab Size} & \textbf{Year} \\
\midrule
GPT-2         &  50{,}257  & 2019 \\
GPT-4         & 100{,}263  & 2023 \\
gpt-oss       & 200{,}019  & 2025 \\
Llama~3.1     & 128{,}256  & 2024 \\
Gemma~3       & 262{,}145  & 2025 \\
Qwen3         & 151{,}669  & 2025 \\
Kimi~K2       & 163{,}842  & 2025 \\
\bottomrule
\end{tabular}
\caption{Tokenizers evaluated in \textsc{Tokka-Bench}.}
\label{tab:tokenizers}
\end{table}

\subsection{Metrics}
\label{sec:metrics}

Table~\ref{tab:metrics} summarizes our five per-language metrics, all computed from a single tokenization pass over the 2\,MB sample.

\begin{table}[t]
\centering
\small
\begin{tabular}{@{}lp{4.6cm}@{}}
\toprule
\textbf{Metric} & \textbf{Description} \\
\midrule
Bytes/Token & Average UTF-8 bytes per token ($\frac{\text{total\_bytes}}{\text{total\_tokens}}$). Higher = more efficient. \\[2pt]
Unique Tokens & Count of distinct token IDs used. Higher = better vocabulary coverage. \\[2pt]
Subword Fertility & Tokens per linguistic unit ($\frac{\text{total\_tokens}}{\text{total\_units}}$). Lower = closer to 1:1 mapping. \\[2pt]
Word-Split \% & Percentage of units requiring $>$1 token. Lower = better boundary alignment. \\[2pt]
Vocab Composition & Script distribution across decoded vocabulary (aggregated). \\
\bottomrule
\end{tabular}
\caption{Per-language metrics in \textsc{Tokka-Bench}.}
\label{tab:metrics}
\end{table}

\paragraph{Cross-linguistic comparability.}
Because UTF-8 byte widths and linguistic-unit definitions vary by script, all metrics are most reliable when comparing tokenizers \emph{within} a single language.
Cross-linguistic comparisons are confounded by encoding overhead and differing unit granularity; see Section~\ref{sec:limitations} for details.

\subsection{Language-Aware Segmentation}
\label{sec:segmentation}

Fertility and word-split metrics require defining a ``linguistic unit.''
Naively applying whitespace segmentation to all languages would produce meaningless fertility values for scripts like Chinese (where words are not space-delimited) or Thai (where spaces separate phrases, not words).
We therefore use three strategies:
\begin{itemize}
\item \textbf{Whitespace segmentation} for alphabetic scripts (Latin, Cyrillic, Arabic, Devanagari, etc.): units are space-delimited words.
\item \textbf{Character segmentation} for logographic and abugida scripts (Chinese, Japanese, Thai, Khmer, Lao, Myanmar): each non-whitespace character is one unit.
\item \textbf{Syllable segmentation} for Tibetan-family scripts: units are tsheg-delimited syllables, with a character-based fallback.
\end{itemize}

\section{Results}

All comparisons below are within-language; cross-linguistic differences are confounded as noted in Section~\ref{sec:metrics}.

\subsection{Tokenization Efficiency}

\begin{figure}[t]
\centering
\includegraphics[width=\columnwidth]{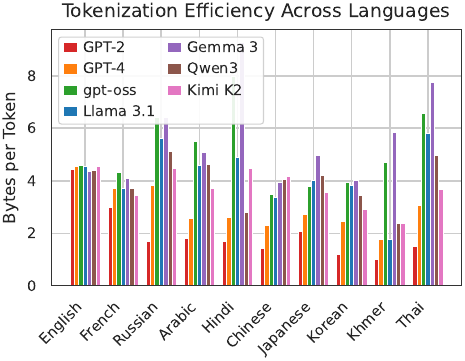}
\caption{Tokenization efficiency (bytes per token) across 10 languages.}
\label{fig:efficiency}
\end{figure}

Figure~\ref{fig:efficiency} shows bytes-per-token across 10 typologically diverse languages.

Clear generational patterns emerge.
GPT-2 consistently trails modern tokenizers, often dramatically: for Hindi, Gemma~3 achieves 9.3 bytes/token versus GPT-2's 1.7 (5.5$\times$), and for Khmer the gap reaches 5.7$\times$.
Even for Russian, modern tokenizers achieve 2.6--3.8$\times$ higher efficiency.
The gains are smallest for English ($<$1.1$\times$), reflecting GPT-2's English-centric vocabulary.
Among recent models, Kimi~K2 and Qwen3 show notably strong performance on Chinese, consistent with CJK-heavy training data, while Gemma~3's larger vocabulary does not uniformly translate to higher efficiency.

\subsection{Vocabulary Coverage}

\begin{figure}[t]
\centering
\includegraphics[width=\columnwidth]{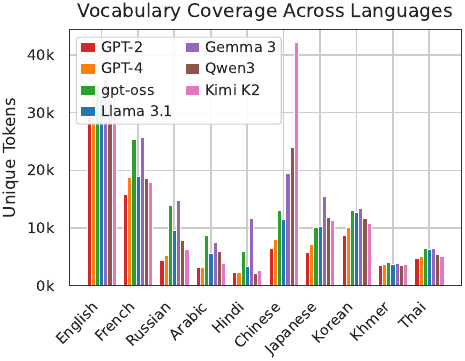}
\caption{Unique token count when encoding 2\,MB per language.}
\label{fig:coverage}
\end{figure}

Figure~\ref{fig:coverage} plots unique token counts.
Within each language, coverage reflects how much vocabulary has been allocated to that script.

Kimi~K2 stands out with exceptionally high Chinese coverage (42K unique tokens vs.\ 6.5K for GPT-2), reflecting dedicated vocabulary allocation.
However, larger vocabularies do not guarantee proportional utilization: averaged across natural languages, GPT-2 activates 16.8\% of its 50K vocabulary per language, while Gemma~3 activates only 5.3\% of its 262K vocabulary---a consequence of spreading capacity across many scripts.

\subsection{Subword Fertility}

\begin{figure*}[t]
\centering
\includegraphics[width=\textwidth]{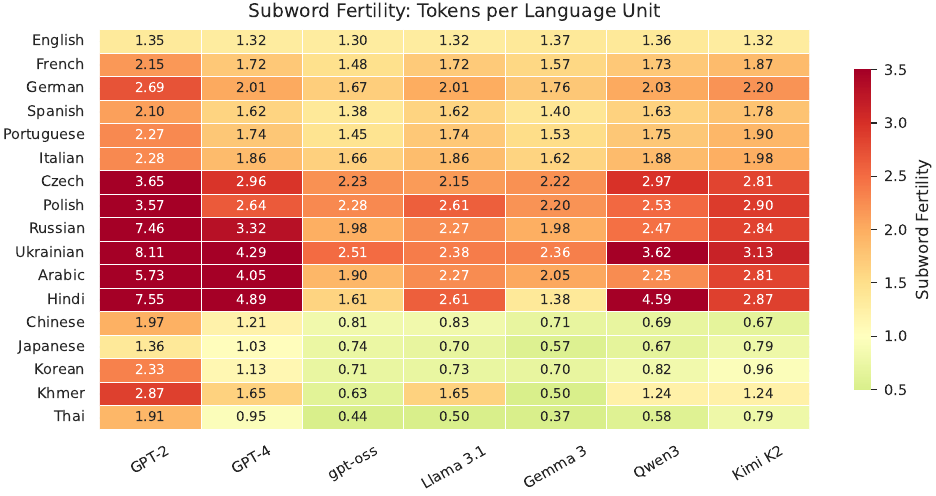}
\caption{Subword fertility (tokens per linguistic unit) across 17 languages and 7 tokenizers. Values near 1.0 are optimal; red cells indicate high fragmentation. Units are language-dependent: words for alphabetic scripts, characters for CJK/Thai/Khmer.}
\label{fig:fertility}
\end{figure*}

Figure~\ref{fig:fertility} presents subword fertility across 17 languages and all seven tokenizers.

Western European languages show low fertility across modern tokenizers (English 1.3, French 1.5, German 1.7 for gpt-oss), while Eastern European and Indic languages exhibit higher fragmentation---Czech and Polish range from 2.2 to 3.7, and GPT-2 reaches 7.5 for Hindi, meaning each word becomes 7--8 tokens on average.
For CJK and Southeast Asian scripts, where units are characters rather than words, several modern tokenizers achieve sub-1.0 fertility (e.g., Gemma~3 at 0.37 for Thai, Kimi~K2 at 0.67 for Chinese), meaning each token spans multiple characters.

\subsection{Programming-Language Convergence}

\begin{figure}[t]
\centering
\includegraphics[width=\columnwidth]{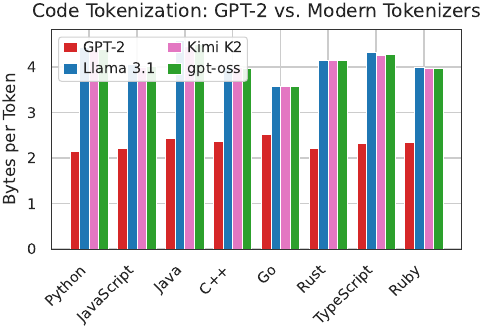}
\caption{Code tokenization efficiency. Modern tokenizers show near-identical performance despite differing natural-language profiles.}
\label{fig:code}
\end{figure}

Figure~\ref{fig:code} compares bytes-per-token for 8 programming languages, contrasting GPT-2 (2019) with three modern tokenizers (Llama~3.1, Kimi~K2, gpt-oss).

Programming-language efficiency has largely converged among post-2023 models: Llama~3.1, Kimi~K2, and gpt-oss achieve nearly identical bytes-per-token despite very different natural-language profiles.
This likely reflects overlapping code corpora and the constrained ASCII character set of programming languages.
GPT-2, trained before large-scale code inclusion, lags by 29--51\%.

\subsection{Vocabulary Size vs.\ Efficiency}

\begin{figure}[t]
\centering
\includegraphics[width=\columnwidth]{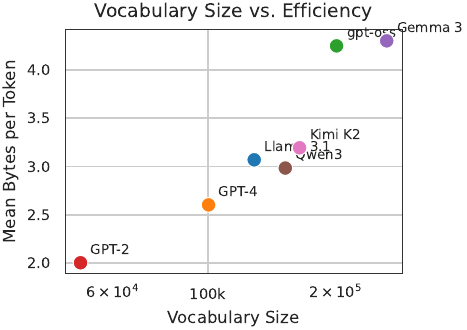}
\caption{Vocabulary size vs.\ mean efficiency across natural languages.}
\label{fig:vocab_eff}
\end{figure}

Figure~\ref{fig:vocab_eff} plots mean bytes-per-token against vocabulary size.

The relationship is noisy: Gemma~3 (262K vocabulary) and gpt-oss (200K) achieve similar mean efficiency (4.30 and 4.25 bytes/token respectively), while Kimi~K2 (164K) achieves 3.19 by concentrating vocabulary on CJK subwords rather than maximizing breadth.
This suggests that \emph{how} vocabulary is allocated matters as much as total size---targeted allocation for high-value scripts can outperform a uniformly larger vocabulary.

\section{Conclusion}

\textsc{Tokka-Bench} provides a reproducible, open-source benchmark for evaluating tokenizers across 100 natural and 20 programming languages with language-aware segmentation, five complementary metrics, and an interactive dashboard.
Our demonstration analysis suggests that vocabulary allocation strategy matters more than raw vocabulary size and that programming-language tokenization has converged among recent models.
The framework and data are available at \url{https://github.com/bgub/tokka-bench}, with an interactive dashboard at \url{https://tokka-bench.streamlit.app/}.

\paragraph{Future work.}
Key directions include (1)~correlating tokenizer metrics with downstream task performance, (2)~developing normalization for fair cross-linguistic comparison, (3)~adding variance estimates via multiple samples per language, and (4)~extending to non-BPE approaches (WordPiece, Unigram).

\section{Limitations}
\label{sec:limitations}

\paragraph{Cross-linguistic comparability.}
Each language is evaluated on a 2\,MB UTF-8 sample.
Because scripts vary in bytes per character (1 for ASCII, 2--3 for Cyrillic/Arabic/Devanagari, 3--4 for CJK), this fixed budget represents different amounts of text across languages, confounding cross-linguistic metric comparisons.
Similarly, bytes-per-token conflates tokenization quality with UTF-8 encoding properties, and fertility uses language-dependent units (words vs.\ characters), making cross-linguistic values incommensurable.
All metrics are most reliable for within-language tokenizer comparisons.

\paragraph{Single sample, no variance estimates.}
Metrics are computed from a single 2\,MB block per language without confidence intervals.
Topic or domain shifts could affect results, particularly unique token counts.

\paragraph{Scope.}
All evaluated tokenizers use BPE variants; the demonstration does not cover WordPiece, Unigram, or byte-level approaches.
We report tokenizer-intrinsic metrics only and do not validate whether they predict downstream task performance.

\section*{Acknowledgments}

Vin Howe, Sachin Raja, and Jacob Holloway reviewed early drafts and provided useful feedback.
Harsha Vardhan Khurdula helped gather relevant research and think through metrics systematically.

\bigskip
\noindent\textit{AI assistance.}
\textit{Claude (Anthropic) was used to assist with drafting, editing, and code generation during the development of this paper and the Tokka-Bench framework.}

\bibliography{references}

\end{document}